# Artificial Intelligence in Reverse Supply Chain Management

The State of the Art

Bo Xing[1*], Wen-Jing Gao[1], Kimberly Battle[1], and Tshilidzi Marwala[1]

[1]Faculty of Engineering and the Built Environment
University of Johannesburg
Johannesburg, South Africa
bxing2009@gmail.com

Fulufhelo V. Nelwamondo[1,2]

[2]Modeling and Digital Science Unit
Council for Scientific & Industrial Research (CSIR)
Pretoria, South Africa

*Abstract*—Product take-back legislation forces manufacturers to bear the costs of collection and disposal of products that have reached the end of their useful lives. In order to reduce these costs, manufacturers can consider reuse, remanufacturing and/or recycling of components as an alternative to disposal. The implementation of such alternatives usually requires an appropriate reverse supply chain management. With the concepts of reverse supply chain are gaining popularity in practice, the use of artificial intelligence approaches in these areas is also becoming popular. As a result, the purpose of this paper is to give an overview of the recent publications concerning the application of artificial intelligence techniques to reverse supply chain with emphasis on certain types of product returns.

**Keywords-artificial intelligence; reverse supply chain management; product returns; end-of-use; end-of-life;**

## I. Introduction

Enterprises around the world are employing reverse supply chain (RSC) practices to overcome the regulations and generate profit making opportunities. As a result of the rapid progress in technology, the product life cycles are shrinking faster than ever. In the face of global competition, heightened environmental regulations and a wealth of additional profits and improved corporate image opportunities, performing the RSC operations at a world class level is becoming essential.

In the literature, different techniques have been employed to solve problems occurring in various dynamic segments of reverse supply chain management (RSCM). Our review is dedicated to the applications of common artificial intelligence (AI) techniques in this context, exploring the current research trends and identifying opportunities for further investigation. The main issues going to be addressed in this article include: what are the main problems within RSCM that have been investigated using AI techniques? What kinds of AI techniques have been employed?

In answer to these questions, we organize our research as follows: first Section II delivers a brief introduction about RSC; next the survey method employed in this research is outlined in Section III; then the results of this survey are detailed in Section IV; finally the discussions and conclusions are given in Section V.

## II. Why Reverse Supply Chain

Reverse supply chains are heralded by environmentalists as key elements of sustainable production. In a forward supply chain, the customer is typically the end of the process. However in changing the end-point of a company's supply chain from the consumer to the product's end-of-life or to possibly even the start of a new production cycle, RSCs present opportunity to extend the use of products, conserve resources, prevent waste, and create secondary markets and jobs in remanufacturing and recycling. Intuitively RSCs are more complicated since return flows may include products, subassemblies and/or materials and may enter the forward supply chain in several return points.

Generally the activities in a RSC vary in complexity and managerial importance from scenario to scenario. The situation is normally complicated by different types of returns [1]:

- Product life cycle returns: These returns are linked to the sales process. The reasons for the returns include problems with products under warranty, damage during transport or product recalls;
- Re-Usable components: These returns are related to consumption, use or distribution of the main product;
- End-of-use returns: These are used products or components that have been returned after customer's usage. These used products are normally traded on an aftermarket or being remanufactured;
- End-of-life returns: These are returns that are taken back from the market to avoid environmental or commercial damage. In the context of take-back laws, these used products are often have to be returned.

Each type of return requires a RSC appropriate to the characteristics of the returned products to optimize value recovery. In this research we will focus on end-of-use (EoU) and end-of-life (EoL) returns. Examples of EoU returns can be seen from waste electrical and electronic equipment (WEEE), which are always discarded for outdate reason. In terms of EoL returns, EU end-of-life vehicles (ELVs) Directive is an instance which aims at reusing and recovering 85% by weight of the average vehicle by the year 2006, and this percentage will increase to 95% by the year 2015 [2]. A simplified material flow in EoU & EoL RSCM is illustrated in Figure 1.

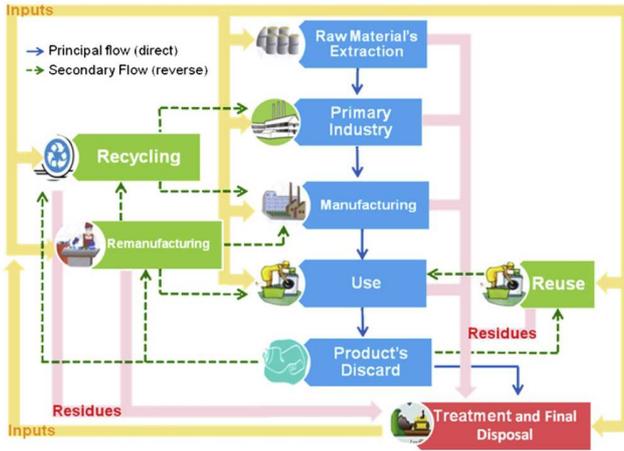

Figure 1. Material flow in EoU & EoL RSCM (adopted from [3]).

As seen from the following figure, the volume of disposal is currently very high. Therefore the objective of RSC strategies for EoU & EoL product returns is to minimize the amount of waste sent to landfills by recovering materials and parts from old or outdated products.

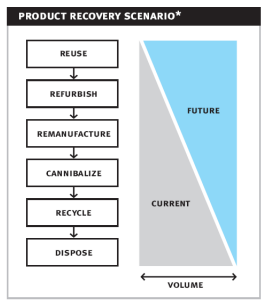

Figure 2. Classification of product recovery scenario (adopted from [4]).

## III. SURVEY METHODOLOGY

The databases used in this study are provided by the library of University of Johannesburg, South Africa. During the research, the following databases have been searched: ScienceDirect, Springer Link, Emerald, and Wiley Interscience. Currently this review covers only journal publications found from aforementioned databases. Meanwhile two groups of keywords (see Table I) are used respectively to cross-search related journal publications in specific online databases.

TABLE I. KEYWORDS FOR THE SURVEY

| Category | Keywords |
|---|---|
| Artificial Intelligence | Ant Colony Optimization (ACO) |
| | Artificial Immune Systems (AIS) |
| | Artificial Neural Network (ANN) |
| | Bayesian Network (BN) |
| | Case Based Reasoning (CBR) |
| | Differential Evolution (DE) |
| | Evolutionary Algorithm (EA) |
| | Evolutionary Programming (EP) |
| | Fuzzy System (FS) |
| | Genetic Algorithm (GA) |
| | Genetic Programming (GP) |
| | Greedy Randomized Adaptive Search Procedure (GRASP) |
| | Memetic Algorithm (MA) |
| | Multi-Agent System (MAS) |
| | Neighborhood Search (NS) |
| | Particle Swarm Optimization (PSO) |
| | Path Relinking (PR) |
| | Reinforcement Learning (RL) |
| | Scatter Search (SS) |
| | Simulated Annealing (SA) |
| | Tabu Search (TS) |
| Reverse Supply Chain Management | Reverse Logistics |
| | End-of-Life Product |
| | Product Returns |
| | Remanufacturing |
| | Product Recovery |

## IV. RESULTS OF LITERATURE SURVEY

The classification of our literature survey results are grouped into the following four aspects:

- AI for RSC network design;
- AI for EoU & EoL products acquisition and assessment;
- AI for EoU & EoL products transportation;
- AI for selection and evaluation of logistics suppliers.

## A. AI for RSC Network Design

To optimize the performance of the RSC, there is a need to establish an effective and efficient infrastructure via optimal network design. As a result, during the past decade, many researchers are paying more and more attention to the network design for product recovery.

Generally RSC network design is concerned with establishing an infrastructure to manage the reverse channel which often consists of final-users, collectors, and remanufacturers. In this category, many AI approaches have been utilized (see Table II). Among various methodologies, GA is the most popular one. An example can be seen from [5] where the authors address the network design problem encountered by an electronic retailer handling consumer returns. A static model is proposed which considers deterministic supply and demand. Then GA is employed to obtain a near optimal solution for the proposed model. The computational experimentation explores the trade-offs between (i) the inventory holding and shipment consolidation costs and (ii) the location costs and customer service (measured in terms of ease of access to collection centers).

Another recent publication in this category using AI approaches is [6], where the authors propose a mixed integer linear programming model to minimize the transportation and fixed opening costs in a multistage reverse logistics network. Due to the NP-hard nature of such network design problems, they apply a SA algorithm with special NS mechanisms to find the near optimal solution.

TABLE II. AI FOR RSC NETWORK DESIGN

| RSCM | AI Approaches | | | | | | | | | |
|---|---|---|---|---|---|---|---|---|---|---|
| | AIS | DE | FS | GA | GRASP | MAS | NS | PSO | SA | TS |
| RSC Network Design | [7] | [8] | [9] [10] [11] [12] [13] [14] | [5] [8] [11] [14] [15] [16] [17] [18] [19] [20] [21] [22] [23] [24] | [24] | [25] | [6] | [19] | [6] [26] | [27] [28] |

## B. AI for EoU & EoL Products Acquisition and Assessment

The condition of the EoU & EoL products acquired by remanufacturing firms through RSC often varies widely. In order to manage this variation, a firm will normally need to collect an over-demanded quantity of used products. By acquiring more excess items, the firm can not only increase its selectivity but also lower the cost of a remanufactured product. Therefore how to obtain these cores in the first place is one of the most challenge issues within remanufacturing industry. However to the best of the authors' knowledge, there is no AI technique has been actually employed in this category. This will definitely be a future research direction.

Meanwhile the determination of the best option for EoU & EoL products is also an important problem faced by companies. There are five commonly used options for EoU & EoL products: direct reuse, repair, remanufacturing, recycling, and disposal. The differences between different options can be summarized as follows:

- The reuse of the whole product as is for its original task is often referred to as direct reuse;
- In repair option, the damaged components of EoU & EoL products are always changed in order to obtain a fully-functional product;
- While the remanufacturing is always considered as the refurbishment of EoU & EoL products up to a quality level similar to a new product;
- In terms of recycling, recovering materials from the EoU & EoL products is always its main objective;
- The landfill or incineration of the EoU & Eol products, which is currently causing serious environment problems, is generally called disposal.

As a result, the development of a decision model to select between these options requires the consideration of various qualitative and quantitative factors such as environmental impact, quality, legislative factors, and cost. The authors of [29] focus on a selection problem of EoL product recovery options for a turbocharger case. The objective is to maximize its recovery value which includes both recovery cost and quality. To solve the problem efficiently, they develop a multi-objective EA algorithm. Other applications of AI techniques in this category are CBR and GA (see Table III).

TABLE III. AI FOR EoU & EoL PRODUCTS ACQUISITION AND ASSESSMENT

| RSCM | AI Approaches | | |
|---|---|---|---|
| | CBR | EA | GA |
| EoU & EoL Products Acquisition and Assessment | [30] | [29] | [31] |

## C. AI for EoU & EoL Products Transportation

From a logistics point of view, EoU & EoL product recovery creates a reverse flow of goods that originates at the locations of product holders, also referred to as customer zones. Once the EoU & EoL products have been consolidated at certain collection center, they will always be transported to disassembly facilities where the disassembly operations, components sorting, components inspection and assessment are normally performed. After this stage, the various components of EoU & EoL products will be delivered to their different final destinations such as remanufacturing plant and disposal sites.

In [32], authors construct a general mixed integer programming model of vehicle routing problem with simultaneous pickups and deliveries and time windows (VRP-SPDTW) cost saving and environmental protection. They propose an improved differential evolution (IDE) algorithm for solving VRP-SPDTW. In the algorithm, a novel decimal coding is first adopted to construct an initial population; then some improved differential evolution operators are used; and in mutation operation, they use an integer order criterion based on natural number coding method. They also introduce a penalty technical to punish the infeasible solution. In addition, in the crossover operation, a self-adapting crossover probability that varied with iteration is also developed. Other AI approaches employed in this category are show in the following table.

TABLE IV. AI IN TRANSPORTATION OF EoU & EoL PRODUCTS

| RSCM | ACO | DE | NS | TS |
|---|---|---|---|---|
| Transportation of EoU & EoL Products | [33] | [32] | [34] | [28] [35] [36] |

### D. AI for Selection and Evaluation of Logistics Suppliers

Logistic suppliers' development is a critical function within EoU & EoL RSCM. Due to the differences between the reverse and forward flows in terms of the cost and complexity of transportation, storage and/or handling operations, many firms outsource their reverse logistic operations to third party logistics providers (3PLs). In [37], a holistic approach based on the ANN and FS is presented for selecting a 3PL in the presence of vagueness. The authors of [38] use fuzzy TOPSIS and interpretive structural modeling for the problem of selection of best 3PL. Similar problem is also addressed in [39] by employing AHP and fuzzy AHP.

TABLE V. AI FOR SELECTION AND EVALUATION OF LOGISTICS SUPPLIERS

| RSCM | AI Approaches | |
|---|---|---|
| | ANN | FS |
| Selection and Evaluation of Logistics Suppliers | [37] | [37] [38] [39] |

## V. DISCUSSIONS AND CONCLUSIONS

As we can see from Table VI, GA and FS are among the most popular AI approaches used in EoU & EoL RSCM. TS is also widely employed in solving EoU & EoL products transportation problems. Based on the review, we can see the complex nature of problems encountered in RSCM always requires multi-objective optimization. Therefore AI seems to be a promising and useful tool to solve these problems and assist practitioner's decision making in a complicated RSC network. But there are some certain aspects in RSCM have not been fully addressed by AI techniques, and also several AI techniques are not employed yet. This is undoubtedly a future direction for both research communities: AI and RSCM.

TABLE VI. LITERATURE ON REVLOG PROCESSES

| AI Approaches | References | No. of References |
|---|---|---|
| Ant Colony Optimization (ACO) | [33] | 1 |
| Artificial Immune Systems (AIS) | [7] | 1 |
| Artificial Neural Network (ANN) | [37] | 1 |
| Case Based Reasoning (CBR) | [30] | 1 |
| Differential Evolution (DE) | [8], [32] | 2 |
| Evolutionary Algorithm (EA) | [29] | 1 |
| Fuzzy System (FS) | [9], [10], [11], [12], [13], [14], [37], [38], [39] | 9 |
| Genetic Algorithm (GA) | [5], [8], [11], [14], [15], [16], [17], [18], [19], [20], [21], [22], [23], [24], [31] | 15 |
| Greedy Randomized Adaptive Search Procedure (GRASP) | [24] | 1 |
| Multi-Agent System (MAS) | [25] | 1 |
| Neighborhood Search (NS) | [6], [34] | 2 |
| Particle Swarm Optimization (PSO) | [19] | 1 |
| Simulated Annealing (SA) | [6], [26] | 2 |
| Tabu Search (TS) | [27], [28], [35], [36] | 4 |

Meanwhile most papers are published in the main stream journals such as Computers & Operations Research, European Journal of Operational Research, and Computers & Industrial Engineering (see Table VII).

TABLE VII. PAPERS PUBLISHED BY JOURNALS

| Computational Intelligence (CI) Approaches | References | No. of References |
|---|---|---|
| Advanced Engineering Informatics | [30] | 1 |
| Applied Mathematical Modeling | [20] | 1 |
| Asia Pacific Journal of Marketing and Logistics | [39] | 1 |
| Computers & Industrial Engineering | [15], [21], [37] | 3 |
| Computers & Operations Research | [8], [16], [18], [23], | 4 |
| European Journal of Operational Research | [7], [28], [36], [11], | 4 |
| Engineering Applications of Artificial Intelligence | [32] | 1 |

| Computational Intelligence (CI) Approaches | References | No. of References |
|---|---|---|
| Environment Science and Technology | [31] | 1 |
| Fuzzy Sets and Systems | [9] | 1 |
| IEEE Transactions on Electronics Packaging Manufacturing | [10] | 1 |
| International Journal of Advanced Manufacturing Technology | [6] | 1 |
| International Journal of Environment Science and Technology | [13] | 1 |
| International Journal of Management and Decision Making | [12] | 1 |
| International Journal of Production Economics | [17], [25] | 2 |
| International Journal of Production Research | [19], [29] | 2 |
| Journal of Global Optimization | [34] | 1 |
| Omega | [5], [24] | 2 |
| Resources, Conservation and Recycling | [22], [38] | 2 |
| Systems Engineering - Theory & Practice | [33] | 1 |
| Transportation Research, Part E | [26], [27] | 2 |
| Transportation Research, Part D | [35] | 1 |
| Transportation Research Record: Journal of the Transportation Research Board | [14] | 1 |

The impetus of RSCM research has come from many aspects, e.g., the recognition of the deleterious effects on the environment of dumping such as electronic wastes, the decreasing capacity of landfills, the compliance with current and forthcoming environmental legislation, and market-driven reasons as well. In this paper, we review various AI methods that have been applied to RSCM. In the future, the developed and developing methods and algorithms should be tested with more practical case studies so as to speed up the implementation progress of RSC.